\documentclass[preprint,12pt]{elsarticle}
\biboptions{super}

\usepackage{amsmath}
\usepackage{booktabs}
\usepackage{xcolor}
\usepackage{colortbl}
\usepackage{graphicx}
\usepackage{lineno}
\usepackage{hyperref}
\usepackage{caption}
\usepackage{adjustbox}
\usepackage{float}
\usepackage{tikz}
\usetikzlibrary{positioning, arrows.meta, calc, shapes.geometric, fit}
\definecolor{oiBlue}{HTML}{0072B2}
\definecolor{oiSky}{HTML}{56B4E9}
\definecolor{oiGreen}{HTML}{009E73}
\definecolor{oiOrange}{HTML}{E69F00}
\definecolor{oiVermilion}{HTML}{D55E00}
\definecolor{oiPurple}{HTML}{CC79A7}
\definecolor{headTeal}{HTML}{265E65}

\journal{Journal of Biomedical Informatics}

\begin{document}

\begin{frontmatter}

\title{Mitigating Hallucinations in Medical Question Answering on ESC
Cardiology Guidelines:\\[2pt] Evaluation and Retrieval Optimization}

\author[hit]{Ilkin Masimov\corref{cor1}}
\ead{ilkinmasimov@outlook.com}
\author[med]{Rufana Baghirli}
\author[hit]{Muhammad Rashid}
\cortext[cor1]{Corresponding author.}

\affiliation[hit]{organization={School of Computer Science and Technology,
Harbin Institute of Technology}, city={Shenzhen}, country={China}}
\affiliation[med]{organization={Harzklinikum Dorothea Christiane Erxleben},
city={Quedlinburg}, country={Germany}}

\begin{abstract}
\textbf{Background:} Large language models (LLMs) produce hallucinations in
medical question answering (QA), and verify-and-revise pipelines that check
atomic claims against retrieved guideline evidence are a standard mitigation.
Published evaluations of such pipelines, however, almost universally treat the
verifier's output as ground truth, an assumption that has rarely been tested.
\textbf{Objective:} To measure, claim by claim against human judgement, how a
verify-and-revise pipeline fails in a guideline-grounded clinical setting, and
to identify which design choices govern its reliability.
\textbf{Methods:} We built a dataset of 160 questions (100
patient vignettes from 89 de-identified real cases, 60 fact-based) over four
recent ESC cardiology guidelines, all reviewed and approved by a practising
internist in two documented rounds, and a retrieval corpus of 627 hand-verified
structured guideline records indexed at two chunk granularities. A five-stage
verify-and-revise pipeline (GPT-5.4 generation, GPT-4.1-mini verification) was
run under four retrieval configurations and three independent runs (1920
arm-runs). All 606 first-pass refuted claim instances were manually audited
with a four-class taxonomy, and every candidate true-hallucination label was
confirmed by the clinical expert; an independent second judge re-judged 906
decisions; and the same expert reviewed 100 pipeline-accepted claims blind.
\textbf{Results:} Only 44.7\%--50.6\% of flagged claims were true
hallucinations: an evaluation trusting the verifier would overstate detected
hallucinations roughly twofold. Correction loops converged in 85.2\%--95.1\% of
cases, yet 35\% of genuine errors in final answers were introduced by
correction itself. Single-run true-hallucination recall was 35.6\%--41.2\%;
the union of three runs reached 71\%--80\%. Two independent verifiers agreed
only moderately (Cohen's $\kappa = 0.505$). The blind probe estimated the
accepted-claim miss rate at 3.2\% (95\% CI 0.7\%--6.7\%). The pipeline reduced
genuine errors in final answers to 1.7\%, at a total project cost of \$51.82.
\textbf{Conclusion:} Claim-level verification reduces hallucinated content by
roughly an order of magnitude, but verifier reliability, rather than generation
quality, is the limiting factor. Verifier precision must be measured, not
assumed; the audit protocol presented here makes that measurement practical.
\end{abstract}


\end{frontmatter}

%\linenumbers  % disabled for arXiv (arXiv forbids line numbers); re-enable for journal review copies

% !Mode:: "TeX:UTF-8"
\section{Introduction}
\label{sec:intro}

Large language models (LLMs) now reach or approach expert-level scores on
examination-style medical benchmarks such as MedQA
\cite{jin2021medqa,singhal2023medpalm,singhal2025medpalm2,nori2023gpt4med},
and they answer free-text clinical questions fluently and apparently
authoritatively. Clinical practice, however, operates on single decisions
rather than average accuracy: a wrong numeric threshold, a wrong drug class,
or a wrong strength of recommendation in a single answer can change a clinical
decision. LLMs are known to produce hallucinations: statements that are
plausible in form but factually wrong or unsupported by any authoritative
source \cite{ji2023hallucination,zhang2023hallucination,huang2025hallucsurvey}, and there are formal arguments that some rate of hallucination is an
innate property of models trained by maximum likelihood on finite data
\cite{xu2024inevitable}. The practical question is therefore not whether
hallucinations can be eliminated, but how often they occur, how reliably they
can be detected, and how much of the damage can be removed after generation.

Clinical practice guidelines are the natural reference standard for studying
these questions, because they encode the current evidence-based consensus in a
precise, graded form: discrete recommendations carrying a class of
recommendation and a level of evidence, together with numeric criteria for
severity, eligibility, and treatment targets. A claim about a guideline can,
at least in principle, be checked objectively against a specific table row.
The guidelines of the European Society of Cardiology (ESC) are a particularly
demanding test bed: they are dense with numeric thresholds and graded
recommendations, and they are updated every few years, so the parametric
knowledge a model absorbed during training is often stale relative to the
current version. We study four recent ESC guidelines: the 2025 ESC/EACTS
guideline on valvular heart disease (VHD) \cite{esc2025vhd}, the 2024 ESC
guideline on atrial fibrillation (AF) \cite{esc2024af}, the 2023 ESC guideline
on cardiomyopathies (CM) \cite{esc2023cm}, and the 2021 ESC guideline on
cardiovascular disease prevention (CVD) \cite{esc2021cvd}. Choosing recent
versions deliberately turns guideline staleness from a confound into a
feature: hallucinations occur naturally and can be studied as they arise.

The standard mitigation for hallucination is retrieval-augmented generation
(RAG) \cite{lewis2020rag,guu2020realm,karpukhin2020dpr,gao2023ragsurvey}. In
the classic retrieve-then-generate arrangement, however, generation and
retrieval are entangled: when the system is wrong, it is difficult to say
which sentence is wrong, which retrieved document was misused, or whether the
error came from retrieval or from generation. The verify-and-revise
architecture \cite{chern2023factool,min2023factscore,wei2024safe,vladika2025}
separates the two concerns: an answer is generated first, decomposed into
atomic claims, each claim is checked against retrieved evidence, and only
refuted claims are rewritten. This produces an explicit audit trail: every
claim carries a verification verdict, and every rewrite is traceable to a
specific refuted claim, which is what makes a claim-level scientific
evaluation possible at all.

What is largely missing for this architecture is an understanding of how it
\emph{fails}, measured claim by claim against human judgement, in a domain
where the errors are quantitative and the reference documents are versioned.
Most published evaluations treat the verifier's output as ground truth: a
claim the verifier flags is counted as an error, and a claim it accepts is
counted as correct. Both assumptions are questionable, and questioning them
requires manual auditing, which is expensive and therefore rare. This paper
asks four research questions:

\begin{itemize}
  \item \textbf{RQ1 (verifier precision).} When the verifier flags a claim as
        refuted by the guideline, how often is that a genuine hallucination
        rather than a verification error, a deviation from an external source,
        or a mismatch between guideline versions?
  \item \textbf{RQ2 (retrieval design).} How do chunk granularity and metadata
        filtering affect what the verifier can catch?
  \item \textbf{RQ3 (correction dynamics).} Do correction loops converge, is
        convergence the same as correctness, and how many errors does
        correction itself introduce?
  \item \textbf{RQ4 (stability and validation).} How stable is the system
        across independent runs of non-deterministic models, and how do its
        verdicts hold up against a second, independent verifier and against
        blind expert review of the claims it accepted?
\end{itemize}

Answering these questions, we make four contributions:

\begin{itemize}
  \item \textbf{C1, claim-level audit with a four-class taxonomy.} All 606
        first-pass refuted claim instances were manually audited and
        classified as true hallucination (284), verifier error (235),
        cross-source deviation (50), or version mismatch (37). Verifier
        precision on true hallucinations is only 44.7\%--50.6\% across
        retrieval arms: a claim-level evaluation that trusts the verifier
        would overstate the detection yield by roughly a factor of two.
  \item \textbf{C2, retrieval design as the studied variable.} Four retrieval
        configurations (coarse chunks, atomic chunks, and atomic chunks with
        guideline-level and section-level metadata filters) are compared on
        equal footing, with shared initial answers across arms to remove the
        generation confound.
  \item \textbf{C3, correction loops are double-edged.} Loops converge in
        85.2\%--95.1\% of cases, but convergence is not correctness: 35\% of
        genuine errors shipped to final answers were introduced by correction
        itself, and 27 of 76 shipped refuted claims are phantom refutations of
        claims that were correct.
  \item \textbf{C4, run instability and second-verifier evidence.}
        Single-run true-hallucination recall is only 35.6\%--41.2\%, while the
        union of three runs reaches 71\%--80\%. Two independent verifiers
        agree only moderately (binary agreement 74.2\%, Cohen's
        $\kappa = 0.505$), and a blind expert probe of 100 pipeline-accepted
        claims estimates the residual miss rate at 3.2\% (95\% CI
        0.7\%--6.7\%).
\end{itemize}

The remainder of the paper is organized as follows.
Section~\ref{sec:related} reviews related work and positions this study.
Section~\ref{sec:data} describes the dataset and retrieval corpus.
Section~\ref{sec:methods} presents the pipeline, the four-arm experimental
design, the audit protocol, and the two validation studies.
Section~\ref{sec:results} reports the results, and
Section~\ref{sec:discussion} discusses implications, limitations, and future
work.

% !Mode:: "TeX:UTF-8"
\section{Related work}
\label{sec:related}

\paragraph{Medical question answering with LLMs}
Medical QA is a standard benchmark domain for LLMs
\cite{jin2021medqa,singhal2023medpalm,singhal2025medpalm2,nori2023gpt4med},
but examination-style accuracy does not measure factual reliability against
current clinical guidelines: the reference standard differs (an answer key
versus a specific, versioned document corpus), and the cost model differs (a
fraction of a percentage point versus a wrong clinical decision). Benchmarking
work on retrieval-augmented medical QA \cite{xiong2024medrag} shows that
retrieval helps but that gains vary widely with corpus and retriever choices,
which foreshadows one of the variables studied here.

\paragraph{Hallucination and its evaluation}
Surveys distinguish intrinsic hallucinations (contradicting a provided source)
from extrinsic ones (unsupported by any source)
\cite{ji2023hallucination,zhang2023hallucination,huang2025hallucsurvey}. Both
matter clinically, but they are not equally checkable: an intrinsic
hallucination can be verified mechanically against the corpus, while an
extrinsic one requires broader evidence or expert judgement. Our audit
taxonomy makes this distinction operational by separating true hallucinations
from cross-source deviations and version mismatches. Detection approaches span
consistency checking \cite{manakul2023selfcheckgpt}, uncertainty estimation,
and external verification; only the last produces an auditable link between an
error and the source it violates. Because clinically meaningful errors occur
at the level of individual claims, claim-level evaluation
\cite{min2023factscore,wei2024safe} resolves what answer-level metrics
conflate; we follow the claim-level approach and extend it with a manual audit
of the verifier's own decisions.

\paragraph{Retrieval-augmented generation}
The effectiveness of RAG \cite{lewis2020rag,guu2020realm,karpukhin2020dpr}
depends heavily on indexing design \cite{gao2023ragsurvey}: chunk granularity,
embedding model, and metadata filtering all affect which evidence reaches the
generator or the verifier. Dense retrieval with small embedding models and
exact flat indexes remains a strong, reproducible baseline
\cite{johnson2019faiss,xiao2023bge}. In guideline corpora, chunking interacts
with document structure in a specific way: recommendation tables carry
per-recommendation class and level labels that are easy to separate from their
context, so coarse chunks dilute the signal while overly fine chunks lose
necessary context. Adaptive RAG variants \cite{asai2024selfrag,yan2024crag}
critique or correct retrieval at generation time but still fuse retrieval and
generation into one pass, inheriting the auditability problem; the
verify-and-revise architecture is complementary, generating without retrieval
and using retrieval afterwards, claim by claim.

\paragraph{Atomic fact checking and verify-and-revise}
FActScore introduced atomic-claim decomposition for evaluating factual
precision \cite{min2023factscore}; FacTool applied the idea across domains
with tool-augmented checking \cite{chern2023factool}; and SAFE scaled it to
long-form evaluation with a search-grounded verifier \cite{wei2024safe}. The
closest prior work is the oncology study of Vladika et al. \cite{vladika2025},
which applies verify-and-revise to medical QA over cancer guidelines with
clinician review of sampled outputs. We follow the same architectural family
and ask what more can be learned when evaluation is pushed from sampling to
exhaustive auditing. Table~\ref{tab:positioning} contrasts the two studies and
the claim-level evaluation line along the dimensions that matter here.

\begin{table}[htbp]
  \centering
  \caption{Positioning against the closest prior work.}
  \label{tab:positioning}
  \small
  \begin{adjustbox}{max width=\textwidth}
    \begin{tabular}{@{}p{3.2cm}p{3.4cm}p{3.0cm}p{4.6cm}@{}}
    \toprule
    Aspect & FActScore / SAFE \cite{min2023factscore,wei2024safe} &
    Vladika et al. \cite{vladika2025} & This study \\
    \midrule
    Domain & open-domain biographies / long-form text & oncology guidelines &
    cardiology guidelines (4 ESC) \\
    \arrayrulecolor{black!18}\midrule\arrayrulecolor{black}
    
    Architecture & decompose + verify & verify-and-revise &
    verify-and-revise, 5 stages \\
    \arrayrulecolor{black!18}\midrule\arrayrulecolor{black}
    
    Reference corpus & web / retrieved & guideline PDFs & structured guideline
    records \\
    \arrayrulecolor{black!18}\midrule\arrayrulecolor{black}
    
    Retrieval studied & no & no & yes, 4 arms (granularity $\times$ filters) \\
    \arrayrulecolor{black!18}\midrule\arrayrulecolor{black}
    
    Claim-level audit & sampled & sampled & exhaustive (606 instances) \\
    \arrayrulecolor{black!18}\midrule\arrayrulecolor{black}
    
    Correction harm measured & no & no & yes (17 instances, 27 phantoms) \\
    \arrayrulecolor{black!18}\midrule\arrayrulecolor{black}
    
    Multi-run stability & no & no & yes (3 runs, union analysis) \\
    \arrayrulecolor{black!18}\midrule\arrayrulecolor{black}
    
    Verifier validation & human spot-check & clinician sample & second judge
    (906 judgements) + blind expert probe (100 claims) \\
    \arrayrulecolor{black!18}\midrule\arrayrulecolor{black}
    
    \bottomrule
  \end{tabular}
    \end{adjustbox}
\end{table}

\paragraph{Verifier reliability}
Using an LLM as a verifier raises the question of how reliable the verifier
itself is; if it is treated as ground truth, its errors propagate silently
into every downstream metric. Agreement between raters is classically measured
by Cohen's $\kappa$ \cite{cohen1960kappa}, with 0.41--0.60 labelled moderate
agreement \cite{landis1977}, and a small blind re-review of accepted items
estimates the miss rate of automatic accept decisions, in the spirit of the
spot-checks in the FActScore and SAFE evaluations
\cite{min2023factscore,wei2024safe}. We use both instruments and treat
moderate agreement not as a defect of the evaluation but as a measurement of
the reliability limits of single-model verification.

% !Mode:: "TeX:UTF-8"
\section{Materials: dataset and retrieval corpus}
\label{sec:data}

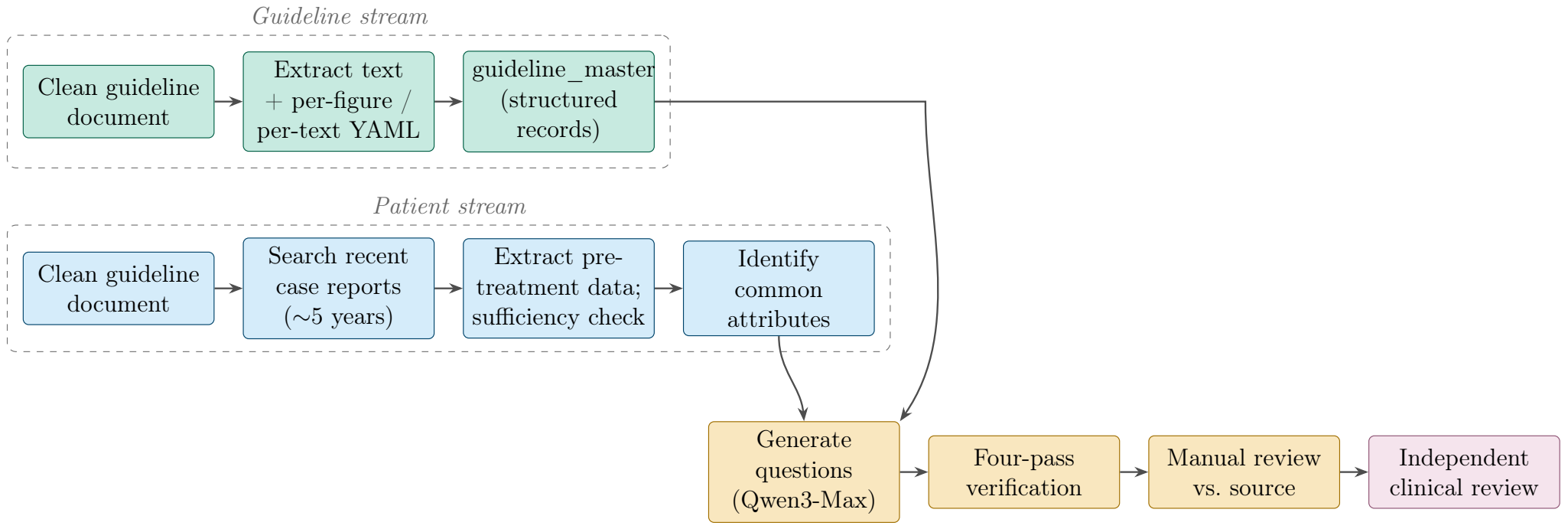
\begin{figure}[htbp]
  \centering
\resizebox{0.98\textwidth}{!}{%
  \begin{tikzpicture}[
      font=\small,
      gbox/.style={rounded corners=2.5pt, fill=oiGreen!22, draw=oiGreen!60!black,
                   align=center, text width=2.75cm, minimum height=1.15cm, inner sep=4pt},
      pbox/.style={gbox, fill=oiSky!25, draw=oiBlue!60!black},
      qbox/.style={gbox, fill=oiOrange!25, draw=oiOrange!70!black},
      rbox/.style={gbox, fill=oiPurple!20, draw=oiPurple!70!black},
      cont/.style={draw=black!55, dashed, rounded corners=4pt, inner sep=7pt},
      arr/.style={-{Stealth[length=2.2mm]}, thick, black!70},
      slab/.style={font=\small\itshape, text=black!60, align=center},
    ]
    % guideline stream
    \node[gbox] (g1) {Clean guideline document};
    \node[gbox, right=0.45cm of g1] (g2) {Extract text + per-figure / per-text YAML};
    \node[gbox, right=0.45cm of g2] (g3) {guideline\_master (structured records)};
    \node[cont, fit=(g1)(g3)] (gcont) {};
    \node[slab, above=1pt of gcont.north] {Guideline stream};

    % patient stream
    \node[pbox, below=1.8cm of g1] (p1) {Clean guideline document};
    \node[pbox, right=0.45cm of p1] (p2) {Search recent case reports ($\sim$5 years)};
    \node[pbox, right=0.45cm of p2] (p3) {Extract pre-treatment data; sufficiency check};
    \node[pbox, right=0.45cm of p3] (p4) {Identify common attributes};
    \node[cont, fit=(p1)(p4)] (pcont) {};
    \node[slab, above=1pt of pcont.north] {Patient stream};

    % question generation and review row
    \node[qbox, below=1.35cm of p4, xshift=0.4cm] (q1) {Generate questions (Qwen3-Max)};
    \node[qbox, right=0.45cm of q1] (q2) {Four-pass verification};
    \node[qbox, right=0.45cm of q2] (q3) {Manual review vs.\ source};
    \node[rbox, right=0.45cm of q3] (q4) {Independent clinical review};

    % edges
    \draw[arr] (g1) -- (g2);
    \draw[arr] (g2) -- (g3);
    \draw[arr] (p1) -- (p2);
    \draw[arr] (p2) -- (p3);
    \draw[arr] (p3) -- (p4);
    \draw[arr] (g3.east) -- ++(4.3,0) to[out=-90, in=55, looseness=0.9] (q1.north east);
    \draw[arr] (p4.south) to[out=-90, in=95] (q1.north);
    \draw[arr] (q1) -- (q2);
    \draw[arr] (q2) -- (q3);
    \draw[arr] (q3) -- (q4);
  \end{tikzpicture}}
  \caption{Dataset construction pipeline.}
  \label{fig:dataset}
\end{figure}

Both artefacts on which the study rests, a dataset of 160 expert-reviewed
clinical questions and a retrieval corpus built from the four ESC guidelines,
were built by hand with model assistance rather than harvested
from existing benchmarks, for three reasons: no published dataset covers these
recent guideline versions; benchmarks rarely provide the per-question source
references that a claim-level audit requires; and expert control of dataset
quality avoids transferring the trust problem from the pipeline's answers to
the dataset itself. Figure~\ref{fig:dataset} gives an overview of the
construction pipeline.

\subsection{Source guidelines}
\label{sec:guidelines}

Four ESC guidelines were selected, weighted by recency and content density
(Table~\ref{tab:guidelines}). The 2025 VHD guideline contributes the most
questions (80 of 160) as the most recent and most threshold-dense document.
The 2021 CVD prevention guideline covers more general preventive content
than the three disease-specific guidelines and was published slightly
earlier; its content is comparatively well represented in model training
data, so it contributes only 10 fact-based questions and serves as a control
for parametric contamination. This recency gradient is intentional: it spans
the realistic range in which clinicians ask about both new and established
guidance.

\begin{table}[htbp]
  \centering
  \caption{The four source guidelines and their roles in the study.}
  \label{tab:guidelines}
  \small
  \begin{adjustbox}{max width=\textwidth}
    \begin{tabular}{@{}p{5.6cm}cc p{5.6cm}@{}}
    \toprule
    Guideline & Year & Code & Role in the study \\
    \midrule
    ESC/EACTS Valvular Heart Disease \cite{esc2025vhd} & 2025 & VHD &
    newest and most threshold-dense; primary test bed (80 questions) \\
    \arrayrulecolor{black!18}\midrule\arrayrulecolor{black}
    
    ESC Atrial Fibrillation \cite{esc2024af} & 2024 & AF &
    recent; risk scores and drug management (50 questions) \\
    \arrayrulecolor{black!18}\midrule\arrayrulecolor{black}
    
    ESC Cardiomyopathies \cite{esc2023cm} & 2023 & CM &
    genotype-specific criteria (20 questions) \\
    \arrayrulecolor{black!18}\midrule\arrayrulecolor{black}
    
    ESC CVD Prevention \cite{esc2021cvd} & 2021 & CVD &
    most general content; contamination control (10 questions) \\
    \arrayrulecolor{black!18}\midrule\arrayrulecolor{black}
    
    \bottomrule
  \end{tabular}
    \end{adjustbox}
\end{table}

\subsection{Guideline knowledge-base engineering}
\label{sec:kb}

Sections with dense, checkable clinical content (imaging criteria, severity
grading, intervention indications, risk stratification, pharmacological
management) were selected from each guideline. Prose was extracted with a
two-column-aware PDF extractor; tables, figures, and boxes were hand-encoded
as structured records and verified side by side against the source PDF, with
class of recommendation and level of evidence preserved as separate fields.
This produced 627 structured records across 43 section files. Two properties
of the knowledge base carry the downstream methodology: \emph{fidelity},
because every table and figure was checked against the PDF by hand, a claim
contradicting a corpus record contradicts the guideline rather than an
extraction artifact; and \emph{traceability}, because every record knows its
exact source location, so every retrieved chunk and every detected
contradiction can be traced back to the original guideline.

\subsection{Patient data sourcing for vignettes}
\label{sec:patients}

Most of the dataset (100 of 160 questions, 62.5\%) consists of
patient-vignette questions. Patient data came from published, de-identified
case reports, all indexed in PubMed Central;
only pre-treatment information was used, so that a question tests how the
guideline should be applied rather than revealing the management actually
undertaken. No new patient data were collected and no identifiable information
was used. Extraction followed a model-assisted procedure with manual control
at every step: Qwen3-Max \cite{yang2025qwen3} first elicited the patient
attributes a realistic vignette must specify; matching case reports were then
searched and their pre-treatment portions separated by manual reading;
Qwen3-Max extracted structured patient details, followed by manual correction;
and a final model-assisted step identified the common attributes grounding
question generation. In total, 82 case reports (2019--2026, mostly 2024--2025)
yielded 89 individual patient cases (several reports describe more than one
patient): 48 for VHD, 22 for AF, and 12 for CM.

\subsection{Question generation and quality control}
\label{sec:qgen}

Questions were drafted with Qwen3-Max at temperature 0.35, deliberately from a
different model family than the GPT models used in the pipeline, so that
systematic biases of one family would not be shared by question generation and
answer generation. Two question types were authored: patient-vignette
questions, which ask how the guideline applies to a concrete clinical
situation, and fact-based questions, which probe specific thresholds, classes
of recommendation, and criteria. Each question was designed to elicit an
answer of at least six to seven sentences, a prerequisite for claim-level
verification. Deterministic Jaccard-similarity gates rejected near-duplicate
questions ($J > 0.55$ within the same patient case, $J > 0.45$ across
patients), and a four-pass automatic verifier checked schema validity,
answerability, absence of answer-leaking phrases, and numeric fidelity of the
stem against the curated source records.

All 160 questions were then reviewed and approved by a clinical expert, a
practising internist with more than ten years of experience, in two
documented rounds (32 sent in round 1, 20 approved; round 2 completed the set
to 140 in four batches of 35). In total, 40 of 160 questions carry an expert
correction. A separate audit-stage review covered hallucination labels
(Section~\ref{sec:audit}). The final composition is 100 vignette and 60
fact-based questions (Table~\ref{tab:dataset}). Every question carries an
identifier encoding its guideline and section group, which later serves as the
join key for the filtered retrieval arms.

\begin{table}[htbp]
  \centering
  \caption{Dataset composition by guideline and question type.}
  \label{tab:dataset}
  \begin{adjustbox}{max width=\textwidth}
    \begin{tabular}{@{}lrrr@{}}
    \toprule
    Guideline & Vignette & Fact-based & Total \\
    \midrule
    2025 ESC/EACTS Valvular Heart Disease (VHD) & 55 & 25 & 80 \\
    2024 ESC Atrial Fibrillation (AF) & 32 & 18 & 50 \\
    2023 ESC Cardiomyopathies (CM) & 13 & 7 & 20 \\
    2021 ESC CVD Prevention (CVD) & 0 & 10 & 10 \\
    \midrule
    \textbf{Total} & \textbf{100} & \textbf{60} & \textbf{160} \\
    \bottomrule
  \end{tabular}
    \end{adjustbox}
\end{table}

\subsection{Retrieval corpus}
\label{sec:corpus}

\begin{figure}[htbp]
  \centering
\resizebox{0.98\textwidth}{!}{%
  \begin{tikzpicture}[
      font=\small,
      hdr/.style={fill=headTeal, text=white, font=\small\bfseries, align=center,
                  text width=4.0cm, inner sep=4pt},
      cell/.style={draw=black!35, fill=white, align=center, minimum width=4.2cm, inner sep=4pt},
      cbox/.style={rounded corners=2.5pt, fill=oiOrange!25, draw=oiOrange!70!black,
                   align=center, text width=4.0cm, minimum height=1.15cm, inner sep=4pt},
      chip/.style={rounded corners=2.5pt, fill=oiGreen!22, draw=oiGreen!60!black,
                   align=center, minimum width=3.0cm, inner sep=4pt},
      arr/.style={-{Stealth[length=2.2mm]}, thick, black!70},
      alab/.style={font=\scriptsize\itshape, text=black!65, align=center},
      note/.style={font=\scriptsize\itshape, text=black!65, align=left},
    ]
    % source table
    \node[hdr] (sh) at (0,0) {Source: Recommendation Table};
    \node[cell, below=0pt of sh] (s1) {Rec.\ 1 -- Class I};
    \node[cell, below=0pt of s1] (s2) {Rec.\ 2 -- Class IIa};
    \node[cell, below=0pt of s2] (s3) {Rec.\ 3 -- Class IIb};
    \node[cell, below=0pt of s3] (s4) {Rec.\ 4 -- Class I};
    \node[cell, below=0pt of s4] (s5) {Rec.\ 5 -- Class III};

    % coarse branch
    \node[cbox, above right=1.0cm and 2.2cm of s1] (coarse)
      {\textbf{COARSE (baseline)}\\ one chunk per table};
    \draw[arr] (s1.east) -- node[alab, above=0pt, sloped, pos=0.5] {whole table} (coarse.west);
    \node[note, right=0.5cm of coarse] {$\rightarrow$ whole table retrieved;\\ the decisive detail is diluted};

    % atomic branch
    \node[chip, below right=1.5cm and 3.0cm of s2] (t1) {Rec.\ 1 -- Class I};
    \node[chip, below=0.25cm of t1] (t2) {Rec.\ 2 -- Class IIa};
    \node[chip, below=0.25cm of t2] (t3) {Rec.\ 3 -- Class IIb};
    \node[chip, below=0.25cm of t3] (t4) {Rec.\ 4 -- Class I};
    \node[draw=black!55, rounded corners=8pt, inner sep=9pt, fit=(t1)(t4)] (atomic) {};
    \node[alab, above=2pt of atomic.north]
      {\textbf{Atomic (optimized)}\\ one chunk per recommendation};
    \draw[arr] (s3.east) -- node[alab, below=1pt, sloped, pos=0.42, xshift=2pt] {per recommendation} (t1.west);
    \node[note, right=0.5cm of t2.south east, anchor=west]
      {$\rightarrow$ single recommendation retrieved;\\ class/level errors exposed directly};
  \end{tikzpicture}}
  \caption{Coarse versus atomic chunking of a recommendation table.}
  \label{fig:chunking}
\end{figure}
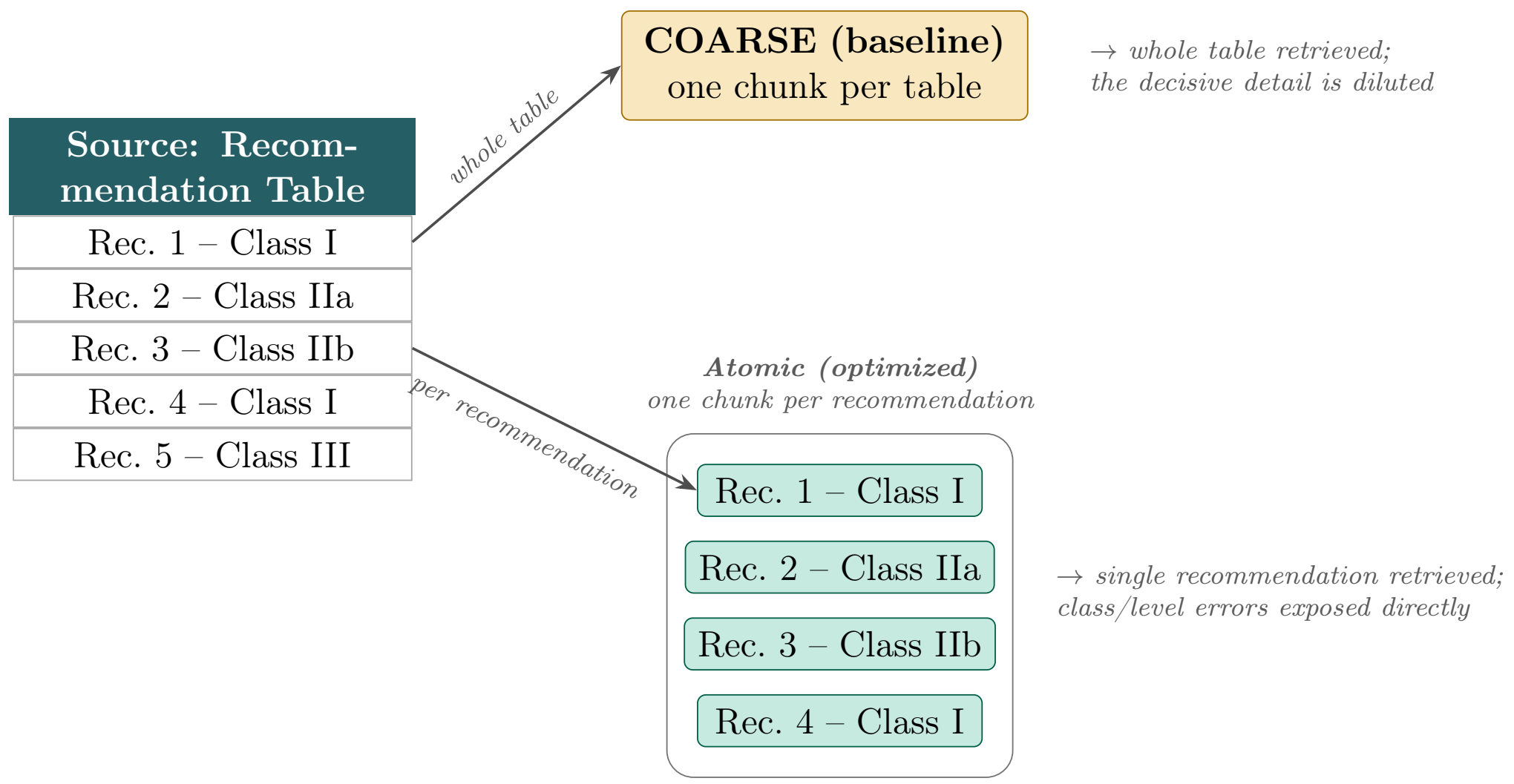

Retrieval supplies the guideline evidence against which claims are checked, so
its design directly determines what the pipeline can detect; we therefore
treat retrieval as a primary object of study rather than fixed infrastructure.
From the 627 structured records, two indexes were built with different chunk
granularities (Table~\ref{tab:corpus}, Figure~\ref{fig:chunking}). The
\emph{coarse} baseline keeps one chunk per record (525 chunks after dropping
boilerplate and empty records). The \emph{atomic} index keeps prose per
subsection, splits recommendation tables per recommendation with class and
level preserved, splits selected lookup tables per row, and keeps figures and
boxes whole (1035 chunks). Both strategies share identical prose handling, so
the comparison isolates structured-content granularity. Every chunk carries
metadata (guideline code, section group, chunk type, and for recommendation
chunks the class and level of evidence) and a position-stable identifier that
makes retrieved evidence traceable across runs. Both indexes use the same
embedding model (bge-small-en-v1.5 \cite{xiao2023bge}) with an exact FAISS
flat inner-product index \cite{johnson2019faiss} on $L_2$-normalised
embeddings (inner product equals cosine similarity), so differences between
arms are attributable to chunking and filtering rather than to the embedder.

\begin{table}[htbp]
  \caption{Corpus composition by chunk type.}
  \label{tab:corpus}
  \begin{tabular}{@{}lrr@{}}
    \toprule
    Chunk type & Coarse & Atomic \\
    \midrule
    prose & 294 & 294 \\
    recommendation & -- & 547 \\
    table row & -- & 82 \\
    table whole & 158 & 39 \\
    figure & 71 & 71 \\
    box & 2 & 2 \\
    \midrule
    \textbf{Total} & \textbf{525} & \textbf{1035} \\
    \bottomrule
  \end{tabular}
\end{table}

% !Mode:: "TeX:UTF-8"
\section{Methods}
\label{sec:methods}

\subsection{The detect-and-correct pipeline}
\label{sec:pipeline}

The pipeline follows the verify-and-revise architecture \cite{vladika2025}: an
answer is generated without retrieval, and retrieval is then used to check and
fix individual claims. Three design principles govern it. First,
\emph{verify-only grounding}: the initial answer is generated without
retrieval, so hallucinations occur naturally and detection and correction can
be studied directly rather than suppressed by retrieval-grounded generation.
Second, an \emph{explicit audit trail}: every claim carries a verification
verdict, a cited chunk, and a rationale, and every rewrite is traceable to
specific refuted claims. Third, \emph{controlled comparison}: a single initial
answer and a single decomposition are produced once per question and verified
under all four retrieval arms, so the only variable across arms is retrieval.

The pipeline has five stages (Figure~\ref{fig:pipeline}):

\begin{figure}[H]
  \centering
\resizebox{0.98\textwidth}{!}{%
  \begin{tikzpicture}[
      font=\small,
      stage/.style={rounded corners=3pt, fill=oiBlue!20, draw=oiBlue!85!black,
                    align=center, text width=3.7cm, inner sep=6pt, minimum height=1.05cm},
      greenbox/.style={stage, fill=oiGreen!25, draw=oiGreen!60!black},
      redbox/.style={stage, fill=oiVermilion!20, draw=oiVermilion!70!black},
      arm/.style={rounded corners=2pt, align=center, text width=2.85cm, inner sep=4pt,
                  minimum height=1.25cm, font=\small},
      arr/.style={-{Stealth[length=2.4mm]}, thick, black!75},
      darr/.style={-{Stealth[length=2.4mm]}, thick, black!70, densely dashed},
      lab/.style={font=\scriptsize, text=black!70, align=center},
    ]
    % top flow
    \node[greenbox] (q) {\textbf{Question}\\(no retrieval)};
    \node[stage, below=0.5cm of q] (s1) {\textbf{Stage 1: Initial Answer}\\GPT-5.4 (generated once)};
    \node[stage, below=0.5cm of s1] (s2) {\textbf{Stage 2: Decompose} atomic claims; tag guideline-fact / patient};

    % stage 3: four retrieval configurations
    \node[redbox, arm] (a1) at ($(s2.south)+(-6.225,-1.6)$) {\textbf{COARSE}\\whole-record baseline};
    \node[greenbox, arm] (a2) at ($(s2.south)+(-2.075,-1.6)$) {\textbf{ATOMIC}\\per-recommendation (class + level)};
    \node[greenbox, arm] (a3) at ($(s2.south)+(2.075,-1.6)$) {\textbf{ATOMIC}\\+ guideline filter};
    \node[greenbox, arm] (a4) at ($(s2.south)+(6.225,-1.6)$) {\textbf{ATOMIC}\\+ section filter};
    \node[draw=black!60, dashed, rounded corners=4pt, inner sep=9pt,
           fit=(a1)(a4), yshift=2pt] (arms) {};
    \node[lab, anchor=west, xshift=6pt, yshift=3pt] at (arms.north west)
      {\textbf{Stage 3:} Retrieve top-$k$ under each configuration};

    % convergence to verification
    \node[stage, text width=6.2cm, below=1.0cm of arms] (s4)
      {\textbf{Stage 4: Verify each claim}\\ supported / refuted / unverifiable\\ (+chunk id, class, rationale)\\ GPT-4.1 mini};
    \node[diamond, aspect=2.1, draw=oiOrange!70!black, fill=oiOrange!28, inner sep=0.5pt,
           below=0.55cm of s4, font=\small, align=center] (d) {any\\refuted};
    \node[stage, below left=1.2cm and 1.0cm of d] (s5) {\textbf{Stage 5: Rewrite refuted claims}\\ GPT-5.4};
    \node[greenbox, below right=1.2cm and 1.0cm of d] (fin) {\textbf{Final Answer}\\ 0 refuted, or 3 iterations reached};

    % edges
    \draw[arr] (q) -- (s1);
    \draw[arr] (s1) -- (s2);
    \draw[arr] (s2.south) -- (arms.north);
    \draw[arr] (a1.south) -- (s4.north west);
    \draw[arr] (a2.south) -- (s4.north -| a2.south);
    \draw[arr] (a3.south) -- (s4.north -| a3.south);
    \draw[arr] (a4.south) -- (s4.north east);
    \draw[arr] (s4) -- (d);
    \draw[arr] (d) -- node[lab, pos=0.4, above left=-4pt] {yes\\($<$ 3 iterations)} (s5);
    \draw[arr] (d) -- node[lab, pos=0.4, above right=-4pt] {no} (fin);
    \draw[darr] (s5.north) to[out=95, in=205, looseness=1]
      node[lab, pos=0.3, left=3pt] {re-verify} (s4.west);
    \end{tikzpicture}}
  \caption{The five-stage detect-and-correct pipeline.}
  \label{fig:pipeline}
\end{figure}
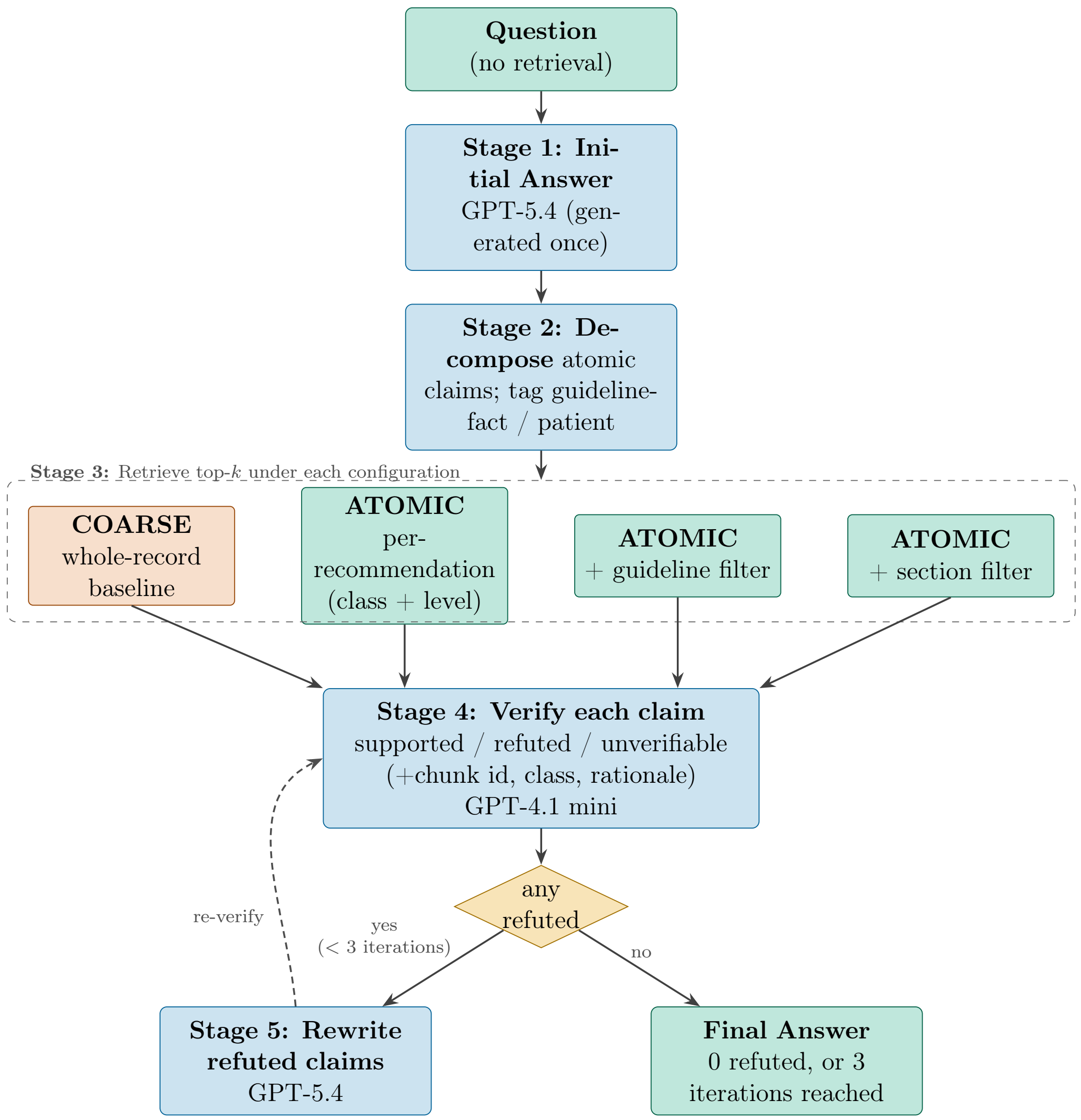

\begin{enumerate}
  \item \textbf{Generation.} An initial answer is produced from the question
        alone, instructed to be focused (six to seven sentences), specific
        about thresholds and recommendation classes, and without hedging,
        since hedged language can smuggle errors past verification. The
        generation prompt asks for guideline-specific answers from the
        model's parametric knowledge only; no guideline text is provided at
        this stage (the full prompts are given in
        Appendix~\ref{app:prompts}).
  \item \textbf{Decomposition.} The answer is decomposed into atomic claims,
        each tagged as \texttt{guideline\_fact} or \texttt{patient\_specific};
        the distinction prevents patient data, which cannot be checked against
        a general guideline, from being incorrectly flagged.
  \item \textbf{Retrieval.} For \texttt{guideline\_fact} claims, the top-$k$
        chunks ($k = 5$) are retrieved by cosine similarity, optionally
        restricted by metadata filters (Section~\ref{sec:arms}). Because the
        index cannot filter on metadata, filtered arms over-fetch 100
        candidates and keep the first $k$ survivors.
  \item \textbf{Verification.} Each claim is judged \emph{supported},
        \emph{refuted}, or \emph{unverifiable}, with a cited chunk identifier
        and a one-line rationale. \texttt{patient\_specific} claims are
        automatically unverifiable; for fact-based questions all claims are
        forced to \texttt{guideline\_fact}.
  \item \textbf{Correction.} Refuted claims are rewritten with up to three
        correction iterations; the loop stops early when no claim is refuted,
        and the rewritten answer is re-decomposed and re-verified each
        iteration.
\end{enumerate}

Generation and rewriting use GPT-5.4 (the stronger model, producing the
consequential clinical text, once per question); decomposition and
verification use GPT-4.1-mini (running once per claim, these stages dominate
the token volume, over 63 million input tokens in the full experiment, and
verification is a constrained classification-style task at roughly one
twentieth of the cost). The APIs are not fully deterministic, which motivates
the multi-run design below.

\subsection{Retrieval arms}
\label{sec:arms}

The experiment covers all 160 questions under four retrieval arms
(Table~\ref{tab:arms}). The arm ladder separates two effects often confounded
in RAG studies: coarse versus atomic isolates chunk granularity at fixed
(absent) filtering, and atomic versus the filtered arms isolates metadata
filtering at fixed granularity. The filtered arms are only possible because of
the traceability built during dataset construction: each question references
the exact guideline sections it was generated from, mirroring the realistic
setting in which a clinician asks a question in the context of a known
guideline and section.

\begin{table}[htbp]
  \centering
  \caption{The four retrieval arms.}
  \label{tab:arms}
  \begin{adjustbox}{max width=\textwidth}
    \begin{tabular}{@{}llll@{}}
    \toprule
    Arm & Index & Chunks & Metadata filter \\
    \midrule
    coarse & coarse & 525 & none \\
    atomic & atomic & 1035 & none \\
    atomic+guideline & atomic & 1035 & guideline code \\
    atomic+guideline+section & atomic & 1035 & guideline code and section group \\
    \bottomrule
  \end{tabular}
    \end{adjustbox}
\end{table}

\subsection{Experimental design}
\label{sec:design}

Each arm is run three times independently, giving $160 \times 4 \times 3 =
1920$ arm-runs. The multi-run design responds directly to API non-determinism:
a single run would report an unstable point estimate of an inherently
stochastic process, whereas three runs allow run-level instability to be
measured. Within a question and a run, one initial answer and one
decomposition serve all four arms, so first-pass verification is directly
comparable across arms; later iterations diverge legitimately because each arm
rewrites and re-decomposes its own answer. All 1920 arm-runs completed without
an unrecoverable failure.

\subsection{Claim-level audit protocol}
\label{sec:audit}

The audit is the methodological centre of the study: it converts the
verifier's refutations from assumed hallucinations into measured ones. Two
units of analysis are kept apart. A \emph{claim instance} is one occurrence of
a refuted claim in one arm of one run of one question (606 first-pass refuted
instances were audited). An \emph{audit row} groups instances of the same
underlying claim within a question, so that a claim flagged several times is
adjudicated once; the 606 instances collapse into 160 audit rows. Claims were
assigned to rows by a constrained matcher with a digit guard (two claims
differing in any number may not share a row) plus seven documented manual
overrides.

Each audited instance was assigned to one of four classes:

\begin{itemize}
  \item \textbf{true\_hallucination} (284 instances, 59 rows): the claim
        contradicts the corpus;
  \item \textbf{verifier\_error} (235): the claim is consistent with the
        corpus but was refuted;
  \item \textbf{cross\_source} (50): the claim is correct per an external
        source and has no direct corpus contradiction;
  \item \textbf{version\_mismatch} (37): the claim follows a different
        guideline version than the corpus.
\end{itemize}

Only true hallucination counts as the phenomenon the pipeline exists to catch;
verifier error counts against verifier precision; cross-source captures
corpus-coverage limits; and version mismatch captures the expected staleness
effect. All 62 candidate true-hallucination rows were additionally confirmed
by the clinical expert: three rows were reclassified to cross\_source and 59
were confirmed. The confirmed rows were further classified by error type: numeric
threshold (a wrong or shifted numeric cut-off, dose, or boundary),
strength/logic (wrong hedging or direction of a recommendation, such as
\emph{should be considered} stated as \emph{is recommended}), class/level
error (a wrong class of recommendation or level of evidence), population
scope (a correct statement applied to a broader or narrower population than
the guideline specifies), and fabricated attribution (a guideline location
cited for content it does not contain).

\subsection{Evaluation metrics}
\label{sec:metrics}

Let $F$ be the set of first-pass flagged (refuted) instances and $T \subseteq
F$ those audited as true hallucinations. The true-hallucination precision of
the verifier is $\operatorname{Prec}_{\mathrm{TH}} = |T|/|F|$. Recall is
defined over the 59 expert-confirmed true-hallucination rows: for a single run
under an arm, row-level recall is the fraction of the 59 rows flagged at least
once; \emph{union recall} counts a row as caught if any of the three runs
flagged it. The denominator is known only for questions the pipeline flagged
somewhere, because the audit covers flagged questions; the expert probe below
therefore exists to bound the complementary quantity, errors hiding in
accepted claims.

A run \emph{converges} when its final iteration contains no refuted claim;
convergence is a property of the loop, not of correctness. Correction outcomes
are classified per audit row into six classes: \texttt{corrected\_clean}
(the final claim is correct), \texttt{narrowed} (the claim survives in a
weakened, more defensible form), \texttt{correction\_harm} (a previously
correct or differently-wrong claim became wrong through rewriting),
\texttt{resisted\_kept} (the claim was kept unchanged despite refutation),
\texttt{unresolved} (still refuted when the iteration cap was reached), and
\texttt{mixed} (instances of the same row ended in different fates across
arms or runs). The
\emph{shipped} analysis counts claims still refuted in final answers and
separates genuine errors from phantom refutations (claims that were correct
but refuted and rewritten).

Agreement between the pipeline verifier and the second judge is measured by
observed agreement and Cohen's $\kappa = (p_o - p_e)/(1 - p_e)$
\cite{cohen1960kappa}, interpreted following Landis and Koch
\cite{landis1977}, with bootstrap confidence intervals (10{,}000
replications). The expert-probe miss rate is estimated with a stratified
estimator: the probe samples two strata with different sampling fractions (89
claims from the random stratum, population share $W_r = 289/300$; 11 from the
judge-disputed stratum, $W_d = 11/300$), and the population miss rate is
$\hat{p} = W_r \hat{p}_r + W_d \hat{p}_d$ with bootstrap confidence intervals.
A naive pooled estimate would overweight the judge-disputed stratum and
overstate the miss rate roughly threefold.

\subsection{Second-verifier study}
\label{sec:secondjudge}

A second, independent judge re-judged the verification decisions. The judge is
GPT-5.4, deliberately a stronger model than the pipeline verifier and a
different model from it, making the judge independent of the specific failure
modes being evaluated. The judge uses the same verification prompt and the
same retrieved passages, resolved from the stored chunk identifiers, so no
retrieval was rerun. Two sets were judged: all 606 first-pass refuted
instances (compared against the human audit labels), and a stratified sample
of 300 first-pass supported claims, 75 per arm (for binary agreement and
Cohen's $\kappa$), giving 906 judgements in total. Verdicts are binarised as
refuse (refuted) versus accept (supported or unverifiable), matching the
decision the pipeline actually acts on.

\subsection{Blind expert probe of accepted claims}
\label{sec:probe}

The audit covers only flagged claims, so it cannot estimate how many incorrect
claims the pipeline \emph{accepted}. To estimate this miss rate, 100
verifier-accepted claims were reviewed blind by the clinical expert: 89
sampled at random from the 289 claims on which the judge agreed with
acceptance, plus all 11 claims on which the judge disagreed. The expert judged
each claim only for clinical correctness, without seeing model verdicts; the
blind design ensures agreement reflects independent judgement rather than
anchoring. Estimation follows the stratified estimator above.

\subsection{Reproducibility}
\label{sec:repro}

All experimental artefacts are frozen with per-file SHA-256 hashes, and all
analysis tables are computed from the frozen inputs by a single analysis
notebook of record. The dataset, corpus, pipeline code, and analysis notebooks
are released with the paper.

% !Mode:: "TeX:UTF-8"
\section{Results}
\label{sec:results}

All numbers are computed from the frozen run payloads and audit files. The
experiment comprises $160 \times 4 \times 3 = 1920$ arm-runs
(63.2 million input tokens, 2.7 million output tokens); all arm-runs
completed without an unrecoverable failure.

\subsection{Flagged claims and true-hallucination precision}
\label{sec:res-precision}

Table~\ref{tab:precision} summarises first-pass verification. The four arms
flag between 141 and 157 claim instances each, and the section-filtered arm
performs best, with the highest true-hallucination precision (50.6\%) and the
most true-hallucination rows caught (47 of 59). Across arms, only
44.7\%--50.6\% of flagged instances are true hallucinations; the remainder are
verifier errors, cross-source deviations, or version mismatches. Against RQ1,
this is the first central finding: a claim-level evaluation that trusted the
verifier would overstate the number of detected hallucinations by roughly a
factor of two. Against RQ2, atomic chunking alone raises the number of caught
true-hallucination instances relative to coarse (72 versus 63) without raising
precision, because it also surfaces more verifier errors; adding the section
filter raises both precision and row coverage, indicating that the filter
removes precise-but-miscontextualised matches rather than changing what the
chunker exposes.

\begin{table}[htbp]
  \centering
  \caption{First-pass flagged instances and audited true-hallucination (TH)
  yield per arm.}
  \label{tab:precision}
  \begin{adjustbox}{max width=\textwidth}
    \begin{tabular}{@{}lrrrr@{}}
    \toprule
    Arm & Flagged & TH instances & TH precision & TH rows caught (of 59) \\
    \midrule
    coarse & 141 & 63 & 44.7\% & 42 \\
    atomic & 154 & 72 & 46.8\% & 42 \\
    atomic+guideline & 157 & 71 & 45.2\% & 43 \\
    atomic+guideline+section & 154 & 78 & 50.6\% & 47 \\
    \bottomrule
  \end{tabular}
    \end{adjustbox}
\end{table}

\subsection{Audit composition}
\label{sec:res-audit}

The 606 flagged instances decompose into 284 true hallucinations, 235 verifier
errors, 50 cross-source deviations, and 37 version mismatches, covering 160
audit rows (59 true-hallucination rows across 39 questions, 75 verifier-error
rows, 18 cross-source rows, 8 version-mismatch rows). How often a row is
flagged across the four arms is itself informative
(Table~\ref{tab:consensus}): rows flagged by all four arms are true
hallucinations 56.4\% of the time, rows flagged by only one arm only 26\%.
Flag multiplicity is thus a usable confidence signal.

\begin{table}[htbp]
  \centering
  \caption{Audit rows by number of flagging arms, crossed with audit label.}
  \label{tab:consensus}
  \small
  \begin{adjustbox}{max width=\textwidth}
    \begin{tabular}{@{}lrrrrrr@{}}
    \toprule
    Arms flagging & cross\_source & true\_halluc. & verifier\_error &
    version\_mism. & All & TH share \\
    \midrule
    1 & 6 & 13 & 28 & 3 & 50 & 26.0\% \\
    2 & 5 & 8 & 21 & 1 & 35 & 22.9\% \\
    3 & 4 & 7 & 8 & 1 & 20 & 35.0\% \\
    4 & 3 & 31 & 18 & 3 & 55 & 56.4\% \\
    \midrule
    \textbf{All} & \textbf{18} & \textbf{59} & \textbf{75} & \textbf{8} &
    \textbf{160} & 36.9\% \\
    \bottomrule
  \end{tabular}
    \end{adjustbox}
\end{table}

\subsection{Error-type profile of true hallucinations}
\label{sec:res-errortypes}

Table~\ref{tab:errortypes} profiles the 59 expert-confirmed
true-hallucination rows by error type. Two thirds are boundary-level
deviations in numeric thresholds (42.4\%) or recommendation strength and logic
(23.7\%) rather than invented content; fabricated attribution, the most
dangerous category clinically, is also the rarest (10.2\%). The hallucinations
that survive to be caught are mostly near-misses: shifted cut-offs, softened
or strengthened recommendation grades, and slightly wrong populations.

\begin{table}[htbp]
  \caption{True-hallucination rows by error type.}
  \label{tab:errortypes}
  \begin{tabular}{@{}lrr@{}}
    \toprule
    Error type & Rows & Share of 59 \\
    \midrule
    numeric threshold & 25 & 42.4\% \\
    strength/logic & 14 & 23.7\% \\
    class/level error & 8 & 13.6\% \\
    population scope & 6 & 10.2\% \\
    fabricated attribution & 6 & 10.2\% \\
    \midrule
    \textbf{Total} & \textbf{59} & 100\% \\
    \bottomrule
  \end{tabular}
\end{table}

\subsection{Run instability and multi-run union}
\label{sec:res-stability}

Single-run true-hallucination row recall is 35.6\%--41.2\%: any individual run
misses a majority of the hallucinations the audit later confirms. The union of
three independent runs raises recall to 42--47 of 59 rows (71\%--80\%), nearly
doubling single-run recall at no additional design cost.
Table~\ref{tab:stability} shows that only about 12\% of caught rows are
flagged in all three runs, while more than half are flagged in exactly one.
This instability is the empirical signature of API non-determinism, and it
makes run multiplicity a practical recall strategy: three runs cost three
times the compute but recover nearly twice the detections. The same logic
applies across arms: the union of all four arms catches all 59
true-hallucination rows (with the caveat that the audit covered flagged
questions only), meaning the arms are complementary. The intersection across
arms contains 31 rows, the most confidently flagged; a consensus rule
requiring at least three of the four arms gives 50.7\% precision and 64.4\%
recall.

\begin{table}[htbp]
  \centering
  \caption{Stability of true-hallucination row detection across three runs.}
  \label{tab:stability}
  \begin{adjustbox}{max width=\textwidth}
    \begin{tabular}{@{}lrrrrr@{}}
    \toprule
    Arm & TH rows (union) & in 1 run & in 2 runs & in 3 runs & \% in all 3 \\
    \midrule
    coarse & 42 & 26 & 11 & 5 & 11.9\% \\
    atomic & 42 & 22 & 15 & 5 & 11.9\% \\
    atomic+guideline & 43 & 24 & 14 & 5 & 11.6\% \\
    atomic+guideline+section & 47 & 27 & 14 & 6 & 12.8\% \\
    \bottomrule
  \end{tabular}
    \end{adjustbox}
\end{table}

\subsection{Correction outcomes}
\label{sec:res-correction}

Correction loops converge in 86.0\%, 89.5\%, 85.2\%, and 95.1\% of cases
across the four arms, with a mean of 1.30 correction iterations; final answers
are clean of flagged claims in 96.2\%--98.8\% of cases. Convergence deserves a
careful reading: a converged answer is one in which the verifier no longer
refutes any claim, which is not the same as a correct answer; the error may
be fixed, the claim may have been weakened until the verifier stopped
objecting, or the verifier may have failed to refute a still-wrong claim.

Of the 59 true-hallucination rows, 38 (64.4\%) end corrected and clean, 8 are
narrowed, 4 suffer correction-introduced harm, 6 remain unresolved, 1 resists
correction, and 2 are mixed (Table~\ref{tab:outcomes}). Two cells repay
attention: the 15 verifier-error rows ending in correction harm are claims
that were correct, refuted anyway, and rewritten into errors, the purest form
of correction damage, arising from the verifier, not the rewriter; and the 11
cross-source rows ending narrowed are claims that were medically defensible
but unsupported by this corpus, characteristically weakened until the verifier
stops objecting, which is safe but erodes content.

\begin{table}[htbp]
  \centering
  \caption{Correction outcome class by audit label (160 rows).}
  \label{tab:outcomes}
  \small
  \begin{adjustbox}{max width=\textwidth}
    \begin{tabular}{@{}lrrrrr@{}}
    \toprule
    Outcome class & cross\_source & true\_halluc. & verifier\_error &
    version\_mism. & All \\
    \midrule
    corrected\_clean & 3 & 38 & 18 & 6 & 65 \\
    narrowed & 11 & 8 & 17 & 0 & 36 \\
    correction\_harm & 0 & 4 & 15 & 0 & 19 \\
    resisted\_kept & 1 & 1 & 12 & 1 & 15 \\
    unresolved & 0 & 6 & 8 & 1 & 15 \\
    mixed & 3 & 2 & 5 & 0 & 10 \\
    \midrule
    \textbf{All} & \textbf{18} & \textbf{59} & \textbf{75} & \textbf{8} &
    \textbf{160} \\
    \bottomrule
  \end{tabular}
    \end{adjustbox}
\end{table}

In the final shipped answers, 76 refuted claims remain: 49 genuine errors and
27 phantom refutations (claims that were correct but refuted and rewritten).
The genuine errors originate from unresolved true hallucinations (26),
corrections that introduced new errors (17, i.e., 35\% of genuine shipped
errors), version mismatches (3), and new errors (3). Genuine-error final
answers total 32 of 1920 (1.7\%). Starting from 284 audited
true-hallucination instances at first pass, the pipeline therefore reduces
hallucinated content reaching final answers by roughly an order of magnitude.
The phantom refutations are invisible to any evaluation that scores only the
initial answers: more than a third of the refuted claims surviving into final
answers were never wrong.

\subsection{Cost}
\label{sec:res-cost}

The full study cost \$51.82 from provider billing records: \$10.06 for dataset
construction (Qwen3-Max), \$35.42 for the pipeline runs (OpenAI), and \$6.34
for the second-verifier study. The batch phase cost \$0.198 per question and
\$0.0165 per arm-run; at these prices, triplicating runs for the union
strategy adds cents, not dollars, per question.

\subsection{Second-verifier study}
\label{sec:res-step7}

The second judge produced 906 judgements. Two findings are encouraging: the
judge confirms 82.4\% of human-audited true hallucinations (234 of 284), so
the hallucinations the pipeline catches are largely real; and on the
300-claim supported sample, the judge disputes only 11 claims (3.7\%), so the
pipeline's accept decisions are highly stable under re-examination.
Table~\ref{tab:crosstab} crosses the human audit labels with the judge
verdicts on the 606 flagged instances: the judge does not confirm 51.9\% of
human-labelled verifier errors (122 of 235), meaning it often agrees with the
original claim where the pipeline verifier objected. Overall binary agreement
is 74.2\% (95\% CI 71.3\%--77.0\%), with Cohen's $\kappa = 0.505$ (95\% CI
0.456--0.554), moderate on the Landis--Koch scale: the two verifiers agree
well beyond chance, but not enough to treat single-model verification as
ground truth. The judge's confirmation rate of true hallucinations is stable
across arms (78.9\%--84.6\%); its rejection rate of verifier errors is highest
in the section-filtered arm (68.8\%), consistent with that arm's cleaner
evidence. Among 44 correction-harm instances, the judge would not have flagged
45.5\%, indicating that a more reliable verification stage would prevent part
of the correction harm.

\begin{table}[htbp]
  \centering
  \caption{Human audit labels versus second-judge verdicts on the 606 flagged
  instances.}
  \label{tab:crosstab}
  \begin{adjustbox}{max width=\textwidth}
    \begin{tabular}{@{}lrrrr@{}}
    \toprule
    Human label & Judge refuted & Judge supported & Judge unverifiable &
    Total \\
    \midrule
    true\_hallucination & 234 & 13 & 37 & 284 \\
    verifier\_error & 113 & 72 & 50 & 235 \\
    cross\_source & 21 & 2 & 27 & 50 \\
    version\_mismatch & 15 & 8 & 14 & 37 \\
    \midrule
    \textbf{Total} & \textbf{383} & \textbf{95} & \textbf{128} &
    \textbf{606} \\
    \bottomrule
  \end{tabular}
    \end{adjustbox}
\end{table}

\subsection{Blind expert probe of accepted claims}
\label{sec:res-probe}

The blind expert review of 100 verifier-accepted claims found 95 correct and 5
incorrect (Table~\ref{tab:probe}). The stratified estimate gives a population
miss rate of 3.2\% (bootstrap 95\% CI 0.7\%--6.7\%), projecting to an
estimated 452 (95--963) potentially incorrect claims among the 14{,}274
pipeline-accepted claims. All five errors are boundary-level deviations rather
than fabrications: a broadened low-risk exercise eligibility window with added
genotypes; an atrial high-rate episode threshold stated as $\geq$175~bpm where
the guideline specifies $\geq$170~bpm (the same error in two different
answers); an office systolic blood-pressure target excluding 130~mmHg, which
the guideline's range includes; and a Class~IIa recommendation upgraded to
Class~I. No fabricated clinical content was identified among the accepted
claims. Of the five errors, the second judge had refuted three and marked one
as unverifiable, indicating that ensemble verification could intercept the
majority of residual errors.

\begin{table}[htbp]
  \centering
  \caption{Expert probe of 100 verifier-accepted claims: error rates by
  stratum.}
  \label{tab:probe}
  \begin{tabular}{@{}lrrrl@{}}
    \toprule
    Stratum & $n$ & Incorrect & Error rate & 95\% CI \\
    \midrule
    random & 89 & 2 & 2.2\% & 0.6\%--7.8\% \\
    judge-disputed & 11 & 3 & 27.3\% & 9.7\%--56.6\% \\
    \midrule
    \textbf{population (weighted)} & 100 & 5 & \textbf{3.2\%} &
    0.7\%--6.7\% \\
    \bottomrule
  \end{tabular}
\end{table}

The stratum contrast is itself a result: judge-disputed accepted claims are
wrong more than twelve times as often as random ones (27.3\% versus 2.2\%), so
judge disagreement is an informative alarm signal rather than noise, and a
rational place to concentrate human review effort in a deployed system.

% !Mode:: "TeX:UTF-8"
\section{Discussion}
\label{sec:discussion}

\subsection{Principal findings}

The positive core of this study is that detect-and-correct works in a
demanding clinical setting. Starting from 284 audited true-hallucination
instances at first pass, the pipeline ships only 32 genuine-error final
answers (1.7\% of all answers), and the expert probe confirms that 95\% of
accepted claims are clinically correct. The correction machinery converges in
up to 95.1\% of cases, and the second judge confirms that 82.4\% of caught
hallucinations are real.

The analysis also identifies where the remaining headroom lies. Fewer than
half of the verifier's refutations are true hallucinations (precision
44.7\%--50.6\%), and two independent verifiers agree only moderately
($\kappa = 0.505$). Verification reliability, rather than generation quality,
is therefore the factor that bounds further improvement. We frame this as a
finding, not a defect: claim-level verification against dense guideline text
is a hard judgement task. The verifier must decide, on five short retrieved
passages, whether a single sentence contradicts a guideline, collapsing
several hard judgements: relevance of the evidence to the claim's population
and context, equivalence of units and wording, and whether an apparent
contradiction is real or an artifact of decomposition. The error-type taxonomy
shows the last judgement dominates: most refutations concern claims whose
wording shifted a threshold or a scope, exactly what a retrieval-only verifier
cannot see through without careful reading.

\subsection{Retrieval design matters, but in a specific way}

Atomic chunking increases the number of contradictions the verifier can
surface, but the highest true-hallucination precision and recall come from the
atomic index combined with guideline and section metadata filters (50.6\%
precision, 47 of 59 rows). The interpretation is that atomic chunks expose
more candidate evidence, while metadata filters remove
precise-but-miscontextualised matches. The complementarity result sharpens
this picture: the all-arm union catches all 59 confirmed rows, so each
configuration catches something the others miss. Retrieval design in guideline
corpora is therefore less about granularity alone and more about granularity
combined with metadata discipline, which in turn depends on the traceability
built during dataset construction. A RAG system over a guideline corpus
inherits the quality of its document engineering to the same degree that it
inherits the quality of its retriever.

\subsection{The correction loop is double-edged}

The correction loop is where the pipeline most actively helps itself, and
where it most actively hurts itself. On the helping side, 64.4\% of confirmed
true-hallucination rows end corrected and clean. On the hurting side, 35\% of
the genuine errors surviving into final answers were introduced by correction
itself, and 27 of 76 shipped refuted claims are phantom refutations of claims
that were never wrong. The mechanism is instructive: a verifier error hands
the rewriter a constraint the evidence cannot satisfy, because the constraint
itself is wrong. The rewriter, instructed to fix exactly the refuted claims,
can comply and introduce a wrong change, weaken the claim until the verifier
stops objecting, or fail to converge, and the audit shows all three
occurring. The correction loop amplifies verifier error into answer-level
damage, which is why verification reliability, and not rewriting quality, is
the limiting factor.

\subsection{The residual error profile is narrow and fixable}

The residual errors share a narrow, encouraging profile: boundary-level
deviations in numeric thresholds or recommendation-strength classes, not
fabricated clinical content. This suggests that, for this corpus and these
questions, the generator's failure mode is not invention but drift: faithful
reproduction of a remembered value from an older guideline, an adjacent
threshold, or a neighbouring patient class. Drift errors are more amenable to
targeted checking than fabrications: a checker that compares numbers and
recommendation verbs between claim and retrieved evidence would have
intercepted four of the five probe errors.

\subsection{Implications for deployment}

For a system meant to sit between a clinician and a guideline, the results
translate into four operating recommendations: (1)~\emph{triplicate runs where
recall matters}: single-run recall leaves half of the detectable
hallucinations on the table, and three runs cost a few cents per question at
the measured prices; (2)~\emph{route human review to judge-disputed
claims}: judge-disputed accepted claims are wrong more than twelve times as
often as random ones, concentrating scarce expert attention where it pays;
(3)~\emph{do not read convergence as cleanliness}: a converged answer can
still contain unresolved genuine errors and phantom-rewritten correct content;
and (4)~\emph{keep the audit trail}: per-claim verdicts, cited chunks, and
rationales are what made an exhaustive manual audit of a 1920-arm-run
experiment feasible.

\subsection{Limitations}

Several limitations bound the claims of this study. First, the audit is
recall-limited by construction: manual labels exist only for flagged
questions, so the audit cannot by itself measure what the verifier missed; the
expert probe partially closes this gap but covers 100 claims, and its estimate
carries a wide interval (0.7\%--6.7\%). Second, clinical review, audit
confirmation, and the probe were all performed by a single expert, without
inter-annotator agreement; the probe was blind, but a second clinician would
strengthen the ground truth. Third, the pipeline and judge use models from a
single vendor, and no external pipeline baseline is compared; the quantitative
evidence is an internal four-arm ablation, and numeric comparisons with other
papers are not meaningful across datasets. Fourth, parametric contamination
cannot be fully excluded for older clinical knowledge, particularly the 2021
CVD guideline. Fifth, API non-determinism bounds the precision of every
reported metric; three runs mitigate but do not eliminate this. Sixth, the
second judge, while independent of the pipeline verifier, is the same model as
the generator, so shared blind spots between generation and judging cannot be
excluded. Finally, 160 questions over four guidelines is a focused instrument,
not a broad benchmark; the results characterise this corpus and domain and
should not be extrapolated to other specialties without new evaluation.

\subsection{Future work}

Four directions follow directly. \emph{Ensemble verification}: combining the
pipeline verifier and an independent judge with AND-gating; the probe suggests
this would intercept four of the five residual errors at some recall cost, and
the existing dual verdicts on all 906 judged instances make an offline
simulation possible without new experiments. \emph{Numeric-consistency
verification}: a targeted checker for thresholds, ranges, and
recommendation-strength classes, the exact failure profile found by the
error-type analysis and the probe. \emph{Re-examination of taxonomy boundary
cases}: the 113 instances that both verifiers refuted but the audit labelled
verifier errors define where single-model verification goes wrong in its most
self-confident form. And \emph{qualitative case studies}: tracing individual
claims through decomposition, verification, and correction, complementing the
aggregate tables with mechanism-level narratives.

\section{Conclusion}
\label{sec:conclusion}

Claim-level verification with correction loops reduces hallucinated content in
guideline-grounded medical QA by roughly an order of magnitude, from 284
audited true-hallucination instances at first pass to 32 genuine-error final
answers (1.7\%), at a total project cost of about fifty dollars. Retrieval
design (atomic chunking with metadata discipline) and multi-run union are
effective and cheap levers. But fewer than half of the verifier's refutations
are true hallucinations, correction itself introduces a third of the remaining
errors, and single verifiers agree only moderately. Verifier precision must be
measured, not assumed. The audit methodology presented here, exhaustive where
prior work sampled and blind where prior work assumed, is the study's most
transferable contribution: any claim-level verification pipeline can be
evaluated the same way, and every such evaluation will be better than one that
trusts its verifier.

\appendix
% !Mode:: "TeX:UTF-8"
\section{Prompt templates}
\label{app:prompts}

The three load-bearing prompt templates are reproduced verbatim from the
pipeline, so that the experiments can be inspected and replicated.

\paragraph{Generation (GPT-5.4, system)} ``You are a cardiology expert
answering a clinical question using ESC guideline knowledge. Write a focused
answer of 6-7 sentences. Be specific about thresholds, classes, and criteria
where relevant. Do not hedge or add disclaimers.'' For rewrites, the user
message additionally lists each refuted claim with the verifier's rationale
and asks for those specific problems to be fixed while keeping everything
else.

\paragraph{Decomposition (GPT-4.1-mini, system)} ``Decompose the answer into
atomic factual claims. Each claim states ONE fact. Tag each claim:
type='guideline\_fact' -> a general medical/guideline statement (thresholds,
classes, drugs, indications) that could be checked against a guideline.
type='patient\_specific' -> a statement about THIS patient's data or an
inference about this individual; NOT checkable against a general guideline.
Return ONLY a JSON array, no prose, no markdown fences. Each element:
\{'id': int, 'claim': str, 'type': 'guideline\_fact'|'patient\_specific'\}.''

\paragraph{Verification (GPT-4.1-mini, system)} ``You verify a single
clinical claim against retrieved ESC guideline passages. Use ONLY the
passages provided. Decide: 'supported' -> a passage clearly supports the
claim. 'refuted' -> a passage clearly contradicts the claim. 'unverifiable'
-> the passages do not contain enough to judge. Return ONLY JSON, no fences:
\{'verdict': str, 'chunk\_id': str|null, 'rationale': str\}. chunk\_id is the
id of the single most relevant passage (or null if none). rationale is ONE
short sentence.''

\section*{CRediT authorship contribution statement}
\textbf{Ilkin Masimov:} Conceptualization, Methodology, Software, Formal
analysis, Investigation, Data curation, Writing -- original draft,
Visualization.

\textbf{Rufana Baghirli:} Validation, Data curation, Writing -- review \&
editing. Clinical review and approval of the dataset questions,
confirmation of the true-hallucination labels and error-type assignments,
and the blind review of pipeline-accepted claims.

\textbf{Muhammad Rashid:} Methodology, Validation, Writing -- review \&
editing.

\section*{Declaration of competing interest}
% TODO: confirm with all authors.
The authors declare that they have no known competing financial interests or
personal relationships that could have appeared to influence the work reported
in this paper.

\section*{Data availability}
The dataset, retrieval corpus, pipeline code, analysis notebooks, frozen
run payloads, and audit workbooks are publicly available at:
\begin{center}
\url{https://github.com/ilkinmasimov/esc-hallucination-qa}
\end{center}

\bibliographystyle{elsarticle-num}
\bibliography{references}

\end{document}